\documentclass[letterpaper]{article} 
\usepackage{aaai24}  
\usepackage{zou_AAAI}
\usepackage{times}  
\usepackage{helvet}  
\usepackage{courier}  
\usepackage[hyphens]{url}  
\usepackage{graphicx} 
\usepackage{natbib}  
\usepackage{caption} 
\title{Generalization Analysis of Machine Learning Algorithms via the Worst-Case Data-Generating Probability Measure}
\author{
Xinying Zou\textsuperscript{\rm 1}, Samir~M.~Perlaza\textsuperscript{\rm 1, \rm2, \rm3} ,
I\~{n}aki~Esnaola\textsuperscript{\rm2, \rm4},
and Eitan~Altman\textsuperscript{\rm1, \rm5}
}

\affiliations {
\textsuperscript{\rm 1} INRIA, Centre Inria d'Université Côte d'Azur, Sophia Antipolis 06902, France\\
\textsuperscript{\rm 2}Department of Electrical and Computer Engineering, Princeton University, Princeton NJ 08544, USA\\
\textsuperscript{\rm 3}GAATI Laboratory, Université de la Polynésie Française, Faaa 98702, French Polynesia\\
\textsuperscript{\rm 4}Department of Automatic Control and Systems Engineering, University of Sheffield, Sheffield S1 3JD, UK\\
\textsuperscript{\rm 5}Laboratoire d’Informatique d’Avignon, Université d’Avignon, France\\
xinying.zou@inria.fr, samir.perlaza@inria.fr, esnaola@sheffield.ac.uk, eitan.altman@inria.fr
}

\begin{document}

\maketitle
	
\section{Abstract}
In this paper, the worst-case probability measure over the data is introduced as a tool for characterizing the generalization capabilities of machine learning algorithms. 
More specifically, the worst-case probability measure is a Gibbs probability measure and the unique solution to the maximization of the expected loss under a relative entropy constraint with respect to a reference probability measure.
Fundamental generalization metrics, such as the sensitivity of the expected loss, the sensitivity of the empirical risk, and the generalization gap are shown to have closed-form expressions involving the worst-case data-generating probability measure.
Existing results for the Gibbs algorithm, such as characterizing the generalization gap as a sum of mutual information and lautum information, up to a constant factor,  are recovered. 
A novel parallel is established between the worst-case data-generating probability measure and the Gibbs algorithm. Specifically, the Gibbs probability measure is identified as a fundamental commonality of the model space and the data space for machine learning algorithms.

\section{Introduction}

The expected generalization error (GE) is a central workhorse for the analysis of generalization capabilities of machine learning algorithms, see for instance \cite{aminian2021exact, aminian2022tighter, xu2017information,chu2023unified} and \cite{Perlaza-ISIT2023b}.
In a nutshell, the GE characterizes the ability of the learning algorithm to correctly find patterns in datasets that are not available during the training stage. Specifically, it is defined for a fixed training dataset and a specific model instance, as the difference between the population risk induced by the model and the empirical risk with respect to the training dataset.

%
When the choice of models is governed by a stochastic kernel, the expected GE (EGE) is the expectation of the GE with respect to the joint-measure of the models and the datasets. 
Closed-form expressions for the EGE are only known for the Gibbs algorithm in the case in which the reference measure is a probability measure \cite{aminian2021exact}; and for the case in which the reference measure is a $\sigma$-finite measure \cite{InriaRR9454}. 

\subsection{Related Works}
In general, the EGE of machine learning algorithms is characterized by various upper-bounds leveraging different techniques. The metric of mutual information was first proposed in \cite{pmlr-v51-russo16}, further developed in \cite{xu2017information} and combined with chaining methods in \cite{asadi2018chaining, asadi2020chaining} for deriving upper bounds on the EGE. Similar bounds on the EGE were obtained in \cite{bu2020tightening, hellstrom2020generalization,hafez2020conditioning, chu2023unified} and references therein. Other information measures such as the Wasserstein distance \cite{aminian2022tighter,lopez2018generalization, wang2019information}, maximal leakage \cite{issa2019strengthened, esposito2020robust}, mutual $f$-information \cite{masiha2023f}, and Jensen-Shannon divergence \cite{aminian2021jensen} were used for providing upper bounds on EGE as well.
In \cite{duchi2021statistics}, the notion of \textit{closeness} of probability measures with respect to a reference measure in terms of statistical distances was used. Therein, the authors explored the case for which the reference is the empirical measure, which is also studied in this work. Such statistical distance was formulated through f-divergences in \cite{duchi2021statistics}, whereas in this work, the statistical distance is described in terms of relative entropy. However, the objective entailed minimizing the expected loss, while this work provides explicit expressions for the difference between empirical risks, population risk, and generalization gap. For the use of $f$-divergences in these optimization problems, see also \cite{InriaRR9521}, and references therein.

Generalization can also be studied as a local minmax problem as in \cite{lee2018minimax}, in which
generalization bounds were given in terms of empirical risks induced by a worst-case probability measure. The set of candidate probability measures in this work was described in terms of the Wasserstein ambiguity set containing the empirical measure and the ground-truth measure almost surely. The minimax formulation was further studied by establishing a correspondence between the principle of maximum entropy and the minimax approach for decision making in \cite{mazuelas2022generalized}. To circumvent the dependence on the statistical description of the dataset, generalization analyses often rely on approaches that decouple the explicit link of the data-generating measure with the GE by using tools from combinatorics \cite{cherkassky1999model}; probability theory \cite{mcallester2003pac, cullina2018pac,haddouche2020pacbayes}; and information theory \cite{xu2017information, aminian2021exact,russo2019much}. These approaches tend to distill the insight about the GE into coarse statistical descriptions of the dataset-generating measures or features of the hypothesis class that the algorithm aims to learn. 

The main drawback of these analytical approaches is that they provide guarantees that entail worst-case dataset generation analysis but do not identify the data-generating measures that curtail the learning capability of the algorithm.  This, in turn, results in descriptions of the EGE for which the dependence on the training dataset and the selected model is not made evident. 
Recent efforts for highlighting the dependence of generalization capabilities on the training dataset have led to explicit expressions for the expectation of the GE when the models are sampled using the Gibbs algorithm in \cite{Perlaza-ISIT2023b,InriaRR9474}. This line of work opens the door to the study of the worst-case data-generating probability measures and their effect on the GE and EGE, as shown in the following section. 

\subsection{Contributions}
The first contribution consists of a probability measure over the datasets coined \emph{the worst-case data-generating} probability measure. Such a measure maximizes the expectation of the loss, while satisfying that its ``\emph{statistical distance}'' to a given probability measure does not exceed a given threshold. In the following, such a ``statistical distance'' is measured via the KL-divergence, also known as relative entropy.   Interestingly, this choice of ``statistical distance'' leads to the fact that, if the worst-case probability measure exists, then it is a Gibbs probability measure (Theorem~\ref{theworstlemma}) parametrized by the reference measure; the ``statistical distance'' threshold; and the loss function.
The variation of the expectation of the loss when the probability measure changes from the worst-case probability measure to an alternative measure is characterized in terms of ``statistical distances'', also  represented by relative entropies. 
Using this result, the variation of the expectation of the loss when the measure changes from an arbitrary measure to any alternative measure is presented (Theorem~\ref{theoremgengapexpression}). This is an important result as the reference measure and the ``statistical distance'' threshold can be arbitrarily chosen, which leads to useful closed-form expressions for such a variation.

The second contribution leverages the observation that under the assumption that datasets are tuples of independent and identically distributed  datapoints, datasets can be represented by their corresponding types \cite{csiszar1998method}, which are also known as empirical probability measures. Interestingly, the empirical risk induced by a model with respect to a given dataset is proved to be equal to the expectation of the loss with respect to the corresponding type (Lemma~\ref{LemmaMadeOnSunday}).  
This observation, in conjuction with Theorem~\ref{theoremgengapexpression} provides an explicit expression to the difference between two empirical risks induced by the same model on two different datasets. This difference is referred to as the \emph{sensitivity} of the empirical risk to variations on the dataset. Using the same arguments, closed-form expressions in terms of ``statistical distances'' are provided for the generalization gap induced by a given model obtained from a given training dataset. 

The final contribution consists of showing that the expected generalization gap and the doubly-expected generalization gap are strongly connected with the notion of worst-case data-generating  probability measure. 
As a byproduct, an alternative proof to the existing result (see \cite{aminian2021exact} and \cite{InriaRR9454}) providing a closed-form expression for the doubly-expected generalization gap of the Gibbs algorithm in terms of mutual and lautum information is presented. Despite the limitation that this alternative proof relies on the assumption of independent and identically distributed data points, its relevance is significant as it highlights an intriguing connection between the Gibbs algorithm and the worst-case data-generating probability measure.

\subsection{Notation}\label{sectionNotation}

Given a measurable space~$\left( \Omega , \mathscr{F} \right)$, the notation~$\triangle\left(\Omega\right)$ is used to represent the set of probability measures that can be defined over~$\left( \Omega , \mathscr{F} \right)$. Often, when the $\sigma$-algebra~$\mathscr{F}$ is fixed, it is hidden to ease notation. Given a measure~$Q \in \triangle\left( \Omega \right)$, the subset~$\triangle_{Q}\left( \Omega  \right)$ of~$\triangle\left( \Omega\right)$ contains all probability measures that are absolutely continuous with respect to the measure~$Q$. 
Given a second measurable space~$\left( \set{X} , \mathscr{G} \right)$,  the notation~$\triangle\left( \Omega | \set{X}\right)$ is used to represent the set of probability measures defined over~$\left( \Omega , \mathscr{F} \right)$ conditioned on an element of~$\set{X}$. Given two probability measures~$P$ and~$Q$ on the same measurable space, such that~$P$ is absolutely continuous with respect to~$Q$,  the relative entropy of~$P$ with respect to~$Q$ is
\begin{equation}\
\label{EqGKL}
\KL{P}{Q} = \int \frac{\mathrm{d}P}{\mathrm{d}Q}(x)  \log\left( \frac{\mathrm{d}P}{\mathrm{d}Q}(x)\right)  \mathrm{d}Q(x),
\end{equation}
where the function~$\frac{\mathrm{d}P}{\mathrm{d}Q}$ is the Radon-Nikodym derivative of~$P$ with respect to~$Q$.

\section{Problem Formulation}
Let~$\set{M}$,~$\set{X}$ and~$\set{Y}$, with~$\set{M} \subseteq \reals^{d}$ and~$d \in \ints$, be sets of \emph{models}, \emph{patterns}, and \emph{labels}, respectively.  
A pair~$(x,y) \in \mathcal{X} \times \mathcal{Y}$ is referred to as a \emph{labeled pattern} or as a \emph{data point}.
Given~$n$ data points, with~$n \in \ints$,  denoted by~$\left(x_1, y_1 \right)$,~$\left( x_2, y_2\right)$,~$\ldots$,~$\left( x_n, y_n \right)$, a dataset is represented by the tuple
\begin{equation}\label{EqTheDataSet}
	\vect{z} = \big(\left(x_1, y_1 \right), \left(x_2, y_2 \right), \ldots, \left(x_n, y_n \right)\big)  \in \left( \set{X} \times \set{Y} \right)^n.
\end{equation}  

Let the function~$f: \set{M} \times \mathcal{X} \rightarrow \mathcal{Y}$ be such that the label assigned to the pattern~$x$ according to the model~$\vect{\theta} \in \set{M}$ is
\begin{equation}\label{EqTheModel}
	y = f(\vect{\theta}, x).
\end{equation}

Let also the function $\hat{\ell}: \set{Y} \times \set{Y} \rightarrow [0, +\infty]$ be such that given a data point~$(x, y) \in \set{X} \times \set{Y}$, the  loss induced by a model~$\vect{\theta} \in \set{M}$ is~$\hat{\ell}\left( f(\vect{\theta}, x), y \right)$.  
In the following, the loss function~$\hat{\ell}$ is assumed to be nonnegative and  for all~$y \in \set{Y}$, it holds that $\hat{\ell}\left( y , y\right) = 0$.

For ease of notation, let the function $\ell: \set{M}\times\set{X}\times\set{Y} \rightarrow [0,+\infty]$ be such that
\begin{IEEEeqnarray}{rCl}
	\label{Eqloss}
	\loss &=& \hat{\ell}\left( f(\vect{\theta}, x), y \right).
\end{IEEEeqnarray}
The \emph{empirical risk} induced by the model~$\vect{\theta}\in \mathcal{M}$, with respect to the dataset~$\vect{z}$ in~\eqref{EqTheDataSet}, is determined by the  function~$\mathsf{L} : \left( \set{X} \times \set{Y} \right)^n \times \set{M} \rightarrow [0, +\infty ]$, which satisfies  
\begin{IEEEeqnarray}{rCl}
	\label{EqLxy}
	\mathsf{L}(\vect{z},\vect{\theta}) & = & 
	\frac{1}{n}\sum_{i=1}^{n}  \ell\left( \vect{\theta}, x_i, y_i\right),
\end{IEEEeqnarray}
where the functions~$f$ and~$\ell$ are defined in~\eqref{EqTheModel} and~\eqref{Eqloss}. 

Using this notation, the problem of model selection is formulated as an empirical risk minimization (ERM) problem, which consists of the optimization problem:
\begin{equation}\label{EqOriginalOP}
\min_{\vect{\theta} \in \set{M}} \mathsf{L} \left(\vect{z}, \vect{\theta} \right).
\end{equation}

The ERM problem is prone to overfitting since the set of solutions to \eqref{EqOriginalOP} are models selected specifically for the given data set $\vect{z}$ in \eqref{EqTheDataSet}, which limits the generalization capability of the resulting optimal model. One way to compensate for overfitting and adding more stability to the learning algorithm is by adding a regularization term to the optimization problem in \eqref{EqOriginalOP}. Such a regularization term can be represented by a function $R: \set{M} \rightarrow \reals$, which yields the regularized ERM problem
\begin{IEEEeqnarray}{rCl}
	\label{regularized-ERM}
	\min_{\vect{\theta} \in \set{M}} \mathsf{L} \left(\vect{z}, \vect{\theta} \right) + \lambda R\left(\vect{\theta}\right),
\end{IEEEeqnarray}where $\lambda$ is a nonnegative real that acts as a regularization parameter. The regularization function $R$ in \eqref{regularized-ERM} constraints the choice of the model, which can be interpreted as requiring a finite space for the models or limiting the “complexity" of the model \cite{shalev2014understanding}. One common choice for $R$ is $R\left(\vect{\theta}\right) =  \|\vect{\theta} \|_{p}$, with $p\ge 1$. The norm is often used to account for the model complexity. Alternatively, the regularization parameter $\lambda$ determines the weight that regularization carries in the model selection.

The main interest in this work is to study the generalization capability for a given model $\vect{\theta} \in \set{M}$ independently from how such a model is chosen.

\section{An Auxiliary Optimization Problem}\label{SectionAuxiliaryProblem}

This section introduces an optimization problem whose solution is referred to as the worst-case data-generating probability measure. This probability measure, which is conditioned on a given model~$\vect{\theta} \in \set{M}$, is parametrized by a probability measure~$P_{S} \in \triangle\left( \set{X} \times \set{Y} \right)$ and by a positive real~$\gamma$. In a nutshell, the worst-case data-generating probability measure maximizes the expected loss while its relative entropy with respect to~$P_S$ is not larger than~$\gamma$. 
Using this notation, the optimization problem of interest is:
\begin{subequations}\label{theworstproblem}
	\begin{IEEEeqnarray}{CCl}
		\label{EqtheworstproblemA}
		\max_{P\in \Delta_{P_S}(\mathcal{X}\times\mathcal{Y})}&&\int\loss\mathrm{d}P(x,y)\\
		\label{EqtheworstproblemB}
		\mathrm{s.t.} 	&&\KL{P}{P_S} \le \gamma\\
		\label{EqtheworstproblemC}	
		&& \int \mathrm{d}P( x,y ) = 1,
	\end{IEEEeqnarray}
\end{subequations}
where the functions~$f$ and~$\ell$ are defined in~\eqref{EqTheModel} and~\eqref{Eqloss}.

The probability measure~$P_S$ in~\eqref{theworstproblem} can be interpreted as a prior on the probability distribution of the datasets. From this perspective, the search of the worst-case probability measure is performed on the set of all probability measures that are at most at  a ``statistical distance'' smaller than or equal to~$\gamma$ from the measure~$P_S$. Here, such a ``statistical distance'' is measured in terms of the relative entropy. The benefits of the choice of relative entropy become apparent when studying the properties of the solution to the optimization problem in~\eqref{theworstproblem}. The impact of the asymmetry of the relative entropy on this problem is left out of the scope of this work. The interested reader is referred to~\cite{Perlaza-ISIT2023a}.
\subsection{The Solution}

The following theorem characterizes the solution to the optimization problem in~\eqref{theworstproblem} using the function~$\CDFl:\reals\rightarrow \reals$, which satisfies
\begin{equation}
	\label{thefunctionJ}
	\CDFl(t) = \log{\left( \int \exp{\left(  t\loss\right)} \mathrm{d}P_{S}(x,y) \right)},
\end{equation}
with the functions~$f$ and~$\ell$ in~\eqref{EqTheModel} and~\eqref{Eqloss}, respectively.

\begin{theorem}\label{theworstlemma}
	The solution to the optimization problem in~\eqref{theworstproblem}, if it exists, is denoted by~$\worstgeneral$ and satisfies for all~$(x,y)\in \supp P_S$,
	\begin{IEEEeqnarray}{rCl}
		\label{Eqtheworstgeneral}
		\frac{\mathrm{d}\worstgeneral}{\mathrm{d}P_S} (x,y) &&=\exp{\left(\frac{\loss}{\beta} 
			- \CDFlS\left(\frac{1}{\beta}\right)\right)},\IEEEeqnarraynumspace
	\end{IEEEeqnarray}
	where the function~$\CDFlS$ is defined in~\eqref{thefunctionJ} and $\beta > 0$ satisfies
	\begin{IEEEeqnarray}{rCl}
		\label{EqLastLabel}	
		\KL{\worstgeneral}{P_S} = \gamma.
	\end{IEEEeqnarray}
\end{theorem}
\begin{IEEEproof}
	The proof is presented in \cite[Appendix A]{InriaRR9515}.
\end{IEEEproof}
Theorem~\ref{theworstlemma} provides a guarantee on the uniqueness of the solution to the optimization problem in~\eqref{theworstproblem}, whenever it exists. Nonetheless, guarantees for the existence of a solution to \eqref{theworstproblem} are not provided. In the following, it is assumed that the model~$\vect{\theta}$, the real~$\gamma$, and the probability measure~$P_S$ in~\eqref{theworstproblem} are such that a solution exists. Let the set~$\mathcal{J}_{P_S, \vect{\theta}}\subset (0, +\infty)$ be:
\begin{IEEEeqnarray}{rcl}
\label{setCDFl}
\mathcal{J}_{P_S, \vect{\theta}} & \triangleq& \left\lbrace t \in \reals: \CDFl \left(\frac{1}{t} \right)  < +\infty  \right\rbrace.
\end{IEEEeqnarray}
The existence of a solution to the problem in \eqref{theworstproblem} is subject to the condition $\CDFlfunction{P_S} \left(\frac{1}{\beta} \right)  < +\infty$, which involves the model $\vect{\theta}$, the loss function $\ell$ in \eqref{Eqloss}, and the parameters $\beta$ and $P_S$. 
This condition is always satisfied in the case in which the function $\ell$ is bounded almost surely with respect to $P_S$, as shown by the following example. 
\begin{example}\label{Example1}
	Assume that for some model $\vect{\theta} \in \set{M}$, there exists a real $a \in \left( 0, +\infty \right)$ such that
	\begin{IEEEeqnarray}{rcl}
		\label{assumption}
		P_{S} \left( \left\lbrace  (x,y) \in \set{X} \times \set{Y} : \ell(\vect{\theta}, x, y) \leqslant a \right\rbrace \right) = 1,
	\end{IEEEeqnarray}
	where the function $\ell$ is defined in \eqref{Eqloss}. Note that the function $\CDFlfunction{P_S}$ satisfies for all $t \in \reals$, 
	\begin{IEEEeqnarray}{rcl}
		\CDFlfunction{P_S} \left(\frac{1}{t} \right)  & \leqslant & \log{\left( \int \exp{\left(  \frac{a}{t} \right)} \mathrm{d}P(x,y) \right)}\\
		& = &  \frac{a}{t} + \log{\left( \int  \mathrm{d}P(x,y) \right)}\\
		&= & \frac{a}{t} <  +\infty,
	\end{IEEEeqnarray}
	which implies that under the assumption in \eqref{assumption}, the optimization problem in \eqref{theworstproblem} always has a solution. 
\end{example}
In general, if a solution to \eqref{theworstproblem} exists, the measure $\worstgeneral$ in~\eqref{Eqtheworstgeneral} is a Gibbs probability measure \cite{georgii2011gibbs}.  
From this perspective, the function $\CDFlfunction{P_S}$ in~\eqref{Eqtheworstgeneral} is often referred to as the log-partition function~\cite{dembo2009large}. Moreover, the probability measure $P_S$ in~\eqref{theworstproblem} can be interpreted as a prior on the probability distribution of the datasets. 


\subsection{Mutual Absolute Continuity}
When the optimization problem in \eqref{theworstproblem} possesses a solution, i.e., $\beta \in \mathcal{J}_{P_S,\vect{\theta}}$ with $\mathcal{J}_{P_S,\vect{\theta}}$ in \eqref{setCDFl}, the loss $\loss$, with $(x,y)\in \supp P_S$, is finite almost surely with respect to $P_S$. 
\begin{lemma}\label{theoremfiniteloss}
If the problem in \eqref{theworstproblem} has a solution, then
	\begin{IEEEeqnarray}{rCl}
		\label{lossfinite}
		&&\ P_S\left(\Bigl\{(x,y)\in \supp P_S:\loss =+\infty\ \Bigl\}\right)=0, \spnum
	\end{IEEEeqnarray}where the function $\ell$ is in \eqref{Eqloss}.
\end{lemma}

\begin{IEEEproof}
	The proof is presented in \cite[Appendix B]{InriaRR9515}.
\end{IEEEproof}
This observation plays a key role in the proof of the main properties of the measure~$\worstgeneral$  in~\eqref{Eqtheworstgeneral}. Among such properties, an important one is the mutual absolute continuity between $\worstgeneral$ and~$P_{S}$, which is formalized by the following lemma.
\begin{lemma}\label{absolutecontinuity}
	The probability measures~$\worstgeneral$ and~$P_{S}$ in~\eqref{Eqtheworstgeneral} are mutually absolutely continuous.
\end{lemma}
\begin{IEEEproof}
	The proof is presented in \cite[Appendix C]{InriaRR9515}.
\end{IEEEproof}
An immediate consequence of the mutual absolute continuity between the measures~$P_{S}$ and~$\worstgeneral$ in~\eqref{Eqtheworstgeneral} is described by the following lemma.

\begin{lemma}\label{lemmafortwolineresult}
	The probability measures~$P_S$ and~$\worstgeneral$ in~\eqref{Eqtheworstgeneral} satisfy:
	\begin{IEEEeqnarray}{rCl}
		\nonumber
		&&\beta \CDFl\left(\frac{1}{\beta}\right)\\
		\label{equality3}
		&=&\int\loss \mathrm{d}\worstgeneral(x,y) - \beta\KL{\worstgeneral}{P_{S}}\supersqueezeequ \spnum\\
		\label{equality4}
		&=&\int \loss \mathrm{d} P_{S}(x,y) + \beta \KL{P_{S}}{\worstgeneral}, \spnum
	\end{IEEEeqnarray}
	where the functions~$f$ and~$\ell$ are defined in~\eqref{EqTheModel} and~\eqref{Eqloss}, respectively; and the function~$\CDFl$ is in~\eqref{thefunctionJ}.
\end{lemma}
\begin{IEEEproof}
	The proof is presented in \cite[Appendix D]{InriaRR9515}.
\end{IEEEproof}


\section{Analysis of the Expected Loss}\label{sectionanalysisonexpectedloss}

Let the function~$G : \set{M}\times \Delta\left(\set{X}\times\set{Y}\right) \times \Delta\left(\set{X}\times\set{Y}\right) \rightarrow \reals$ be such that 
\begin{IEEEeqnarray}{rCl}
	\label{Eqgengap}
	&&G(\vect{\theta}, P_1, P_2)\\
	\nonumber
	\nonumber
	&&=  \int\loss\mathrm{d}P_{1}(x,y) - \int\loss\mathrm{d}P_{2}(x,y)\middlesqueezeequ,
\end{IEEEeqnarray}
where the functions~$f$ and~$\ell$ are defined in~\eqref{EqTheModel} and~\eqref{Eqloss}, respectively.
The value~$G(\vect{\theta}, P_1, P_2)$ represents the variation of the expectation of the loss when the probability measure over the data points changes from~$P_{2}$ to~$P_1$. 
Such a value is often referred to as the \emph{sensitivity} of the expected loss to variations on the probability distribution of the data points. Such a sensitivity  is characterized by the following theorem for the specific case of variations from the measure~$\worstgeneral$ in~\eqref{Eqtheworstgeneral} to an alternative measure.

\begin{theorem}[Sensitivity of the Expected Loss]\label{deviationfromtheworst}
	For all~$P \in \Delta_{P_{S}} \left(\mathcal{X}\times\mathcal{Y}\right)$ and for all~$\vect{\theta} \in \set{M}$, 
	\begin{IEEEeqnarray}{rCl}
		\label{Eqdeviationworst}
		&&G\left(\vect{\theta}, P, \worstgeneral\right)\\
		\nonumber
		&&=\beta\left(\KL{P}{P_{S}} - \KL{P}{\worstgeneral} -  \KL{\worstgeneral}{P_{S}} \right),
	\end{IEEEeqnarray}
	where the functional~$G$ is defined in~\eqref{Eqgengap}; and the model~$\vect{\theta}$ and the measures~$P_S$ and~$\worstgeneral$ satisfy~\eqref{Eqtheworstgeneral}.
\end{theorem}
\begin{IEEEproof}
	The proof is presented in \cite[Appendix E]{InriaRR9515}.
\end{IEEEproof}

The following corollary of Theorem~\ref{deviationfromtheworst} describes the sensitivity of the expected loss  for variations from~$\worstgeneral$  to the reference measure~$P_S$.

\begin{corollary}\label{lemmagapworstreference}
The probability measures~$P_S$ and~$\worstgeneral$ in~\eqref{Eqtheworstgeneral} satisfy:
	\begin{IEEEeqnarray}{rCl}
	\nonumber
&&G(\vect{\theta}, P_{S}, \worstgeneral)\\
\label{Eqgapworstreference}
		&=&- \beta\left(\KL{P_{S}}{\worstgeneral} + \KL{\worstgeneral}{P_{S}}\right),
	\end{IEEEeqnarray}
	where the functional~$G$ is in~\eqref{Eqgengap}.
\end{corollary}

The right-hand side of the equality in~\eqref{Eqgapworstreference} is a symmetrized Kullback-Liebler divergence, also known as Jeffrey's divergence~\cite{jeffreys1946invariant}, between the measures~$P_S$ and~$\worstgeneral$.
More importantly, it holds that $\KL{P_{S}}{\worstgeneral} \geqslant 0$ and~$\KL{\worstgeneral}{P_{S}} \geqslant 0$, which reveals the fact that the expected loss induced by the Gibbs probability measure~$\worstgeneral$ is larger than or equal to  the expected loss induced by the reference measure~$P_S$. This is formalized by the following corollary of Theorem~\ref{deviationfromtheworst}.

\begin{corollary}\label{CorTheWorstCorollary}The probability measures $P_S$ and $\worstgeneral$ in~\eqref{Eqtheworstgeneral} satisfy:
	\begin{IEEEeqnarray}{rCl}
		\nonumber
		\int\loss \mathrm{d} \worstgeneral (x,y) &\ge& \int\loss \mathrm{d}P_S(x,y),\Tsupersqueezeequ
	\end{IEEEeqnarray}where the function $\ell$ is defined in \eqref{Eqloss}.
\end{corollary}
Note that the probability measure~$P_S$ in Corollary~\ref{CorTheWorstCorollary} can be arbitrarily chosen. That is, independent of the model~$\vect{\theta}$. From this perspective, the measure~$P_S$ can be interpreted as a prior on the datasets, while the probability measure~$\worstgeneral$ can be interpreted as a posterior for the worst-case once the prior~$P_S$ is confronted with the model~$\vect{\theta}$.   

Equipped with the exact characterization of the sensitivity from the measure~$\worstgeneral$ to any alternative measure~$P$ provided by Theorem~\ref{deviationfromtheworst}, it is possible to obtain the sensitivity of the expected loss when the measure changes from a given probability measure to any alternative probability measure, as shown by the following theorem.

\begin{theorem}\label{theoremgengapexpression}
	For all~$P_1 \in \Delta_{P_S} \left(\set{X}\times\set{Y}\right)$ and~$P_2 \in \Delta_{P_S} \left(\set{X}\times\set{Y}\right)$, and for all~$\vect{\theta}\in \set{M}$, the functional~$G$ in~\eqref{Eqgengap} satisfies
	\begin{IEEEeqnarray}{rCl}
		\nonumber
G(\vect{\theta},P_1,P_2) 
		&=& \beta\Bigl(\KL{P_2}{\worstgeneral}-\KL{P_1}{\worstgeneral} \\
		\label{Eqexpressiongengap}
		&&- \KL{P_2}{P_S} + \KL{P_1}{P_S}\Bigl),
	\end{IEEEeqnarray}
	where the model~$\vect{\theta}$ and the measures~$P_S$ and~$\worstgeneral$ satisfy~\eqref{Eqtheworstgeneral}.
\end{theorem}
\begin{IEEEproof}
The proof is presented in \cite[Appendix F]{InriaRR9515}.
\end{IEEEproof}

Note that the parameters~$\gamma$ and~$P_S$ in~\eqref{theworstproblem} can be arbitrarily chosen. This is essentially because only the right-hand side of~\eqref{Eqexpressiongengap} depends on $P_S$ and $\beta$. 
Another interesting observation is that none of the terms in the right-hand side of~\eqref{Eqexpressiongengap} depends simultaneously on both~$P_1$ and~$P_2$. Interestingly, these terms depend exclusively on  the pair formed by~$P_{i}$ and~$P_S$, with~$i \inCountTwo$.
These observations highlight the significant flexibility of the expression in~\eqref{Eqexpressiongengap} to construct closed-form expressions for the sensitivity~$G(\vect{\theta},P_1,P_2)$ in~\eqref{Eqgengap}. 
The only constraint on the choice of~$P_S$ is that both measures~$P_1$ and~$P_{2}$ must be absolutely continuous with respect to~$P_S$.

Two choices of~$P_S$ for which the expression in the right-hand side of~\eqref{Eqexpressiongengap} significantly simplifies are~$P_S = P_1$ and~$P_S = P_2$, which leads to the following corollary of Theorem~\ref{theoremgengapexpression}.

\begin{corollary}\label{CorRGspecifica}
	If~$P_1$ is absolutely continuous with~$P_2$, then the value~$G(\vect{\theta}, P_1, P_2)$ in~\eqref{Eqgengap} satisfies: 
	\begin{IEEEeqnarray}{rcl}
		\nonumber
		&& G(\vect{\theta}, P_1, P_2) \\
		&=&\beta\Bigl(\KL{P_2}{P_{Z|\vect{\Theta} = \vect{\theta}}^{\left( P_2, \beta \right)}}-\KL{P_1}{P_{Z|\vect{\Theta} = \vect{\theta}}^{\left( P_2, \beta \right)}}   + \KL{P_1}{P_2}\Bigl)\supersqueezeequ. \IEEEeqnarraynumspace
	\end{IEEEeqnarray}
	Alternatively, if~$P_2$ is absolutely continuous with~$P_1$ then,  
	\begin{IEEEeqnarray}{rcl}
		\nonumber
		&&  G(\vect{\theta}, P_1, P_2) \\ 
		&=&\beta\Bigl(\KL{P_2}{P_{Z|\vect{\Theta} = \vect{\theta}}^{\left( P_1, \beta \right)}}-\KL{P_1}{P_{Z|\vect{\Theta} = \vect{\theta}}^{\left( P_1, \beta \right)}} -  \KL{P_2}{P_1}\supersqueezeequ, \IEEEeqnarraynumspace
	\end{IEEEeqnarray}
	where for all~$i \inCountTwo$, the probability measure~$P_{Z|\vect{\Theta} = \vect{\theta}}^{\left( P_i, \beta \right)}$ satisfies~\eqref{Eqtheworstgeneral} under the assumption that~$P_S = P_i$.
\end{corollary}

Interestingly, absolute continuity of~$P_1$ with respect to~$P_2$ or of~$P_2$ with respect to~$P_1$ is not necessary for obtaining an expression for the value~$G(\vect{\theta}, P_1, P_2)$ in~\eqref{Eqgengap}. Note that choosing~$P_S$ as a convex combination of~$P_1$ and~$P_2$, guarantees an explicit expression for~$G(\vect{\theta}, P_1, P_2)$ independently of whether these measures are absolutely continuous with respect to each other.

\section{Analysis of the Empirical-Risk}\label{sectionanalysisofempirical}

This section presents a mathematical object known as a \emph{type} in the realm of information theory \cite{csiszar1998method}. In the context of this work, a type is a probability measure induced by a dataset, as shown hereunder. 
\begin{definition}[The Type]\label{DefType}
The type induced by the dataset~$\vect{z}$ in~\eqref{EqTheDataSet} on the measurable space~$\left(\mathcal{X}\times\mathcal{Y}, \mathscr{F}_{\mathcal{X}\times\mathcal{Y}}\right)$, denoted by~$P_{\vect{z}}$, is such that for all singletons~$\lbrace (x,y) \rbrace \in \mathscr{F}_{\mathcal{X}\times\mathcal{Y}}$,
	\begin{IEEEeqnarray}{rCl}
		\label{Eqthetype}
		P_{\vect{z}}\left( \lbrace (x,y) \rbrace \right) &&=\frac{1}{n}\sum_{t=1}^{n}\ind{x=x_t, y=y_t}(x,y).
	\end{IEEEeqnarray}
\end{definition}
This definition illustrates the reason why the type is often referred to as \textit{empirical probability measure}. In the following, the abuse of noting~$P_{\vect{z}}\left( \lbrace (x,y) \rbrace \right)$ as~$P_{\vect{z}}\left(x,y\right)$ is allowed for ease of presentation.
The central observation of this section is that the empirical risk~$\mathsf{L}(\vect{z},\vect{\theta})$ in~\eqref{EqLxy} can be written as the expectation of the loss with respect to the type~$P_{\vect{z}}$. This is formalized by the following lemma.
\begin{lemma}[Empirical Risks and Types]\label{LemmaMadeOnSunday}
The empirical risk~$\mathsf{L}(\vect{z},\vect{\theta})$ in~\eqref{EqLxy} satisfies
\begin{IEEEeqnarray}{rCl}
	\mathsf{L}(\vect{z},\vect{\theta}) &=& \int\loss\mathrm{d}P_{\vect{z}}(x,y),
\end{IEEEeqnarray}
where the measure~$P_{\vect{z}}$ is the type induced by the dataset~$\vect{z}$ in~\eqref{EqTheDataSet}; and the functions~$f$ and~$\ell$ are defined in~\eqref{EqTheModel} and~\eqref{Eqloss}, respectively.
\end{lemma}
\begin{IEEEproof}
	The proof is presented in \cite[Appendix G]{InriaRR9515}.
\end{IEEEproof}

\subsection{Sensitivity of the Empirical Risk}

Equipped with the result in Lemma~\ref{LemmaMadeOnSunday}, for a fixed model, the sensitivity of the empirical risk to changes on the datasets can be characterized using the results obtained in the previous section for the expected loss.
More specifically, consider the two datasets~$\vect{z}_1 \in \left(\set{X} \times \set{Y} \right)^{n_1}$ and~$\vect{z}_2 \in \left(\set{X} \times \set{Y} \right)^{n_2}$ that induce the types~$P_{\vect{z}_1}$ and~$P_{\vect{z}_2}$, respectively. Hence, given a model~$\vect{\theta} \in \set{M}$, it follows that
\begin{IEEEeqnarray}{rCl}
\label{definitionrecall}
G(\vect{\theta}, P_{\vect{z}_1}, P_{\vect{z}_2}) & = &\mathsf{L}(\vect{z}_1,\vect{\theta}) - \mathsf{L}(\vect{z}_2,\vect{\theta}),
\end{IEEEeqnarray}
where the functional~$G$ is in~\eqref{Eqgengap}. 
Assume that~$P_{\vect{z}_1}~$ and~$P_{\vect{z}_2}$ are absolutely continuous with respect to the reference measure~$P_S$ in~\eqref{theworstproblem}. Under this assumption, the equality in~\eqref{definitionrecall} leads to a characterization of the sensitivity of the empirical risk induced by a given model~$\vect{\theta}$ when the dataset is changed from $\vect{z}_1$ to $\vect{z}_2$. 

\begin{theorem}\label{theoremdifferenceL}
	Given two datasets~$\vect{z}_1 \in {\left(\set{X}\times \set{Y}\right)}^{n_1}$ and~$\vect{z}_2 \in {\left(\set{X}\times \set{Y}\right)}^{n_2}$ whose types~$P_{\vect{z}_1}$ and~$P_{\vect{z}_2}$ are absolutely continuous with respect to the measure~$P_S$ in~\eqref{theworstproblem}, the following holds for all~$\vect{\theta} \in \set{M}$:
	\begin{IEEEeqnarray}{rCl}
	\nonumber
&&\mathsf{L}(\vect{z}_1,\vect{\theta}) - \mathsf{L}(\vect{z}_2,\vect{\theta})\\
\nonumber
 &= &\beta \Bigl(\KL{P_{\vect{z}_2}}{\worstgeneral} - \KL{P_{\vect{z}_1}}{\worstgeneral} \\
\label{EqDifferenceLOnTheBeach}
		&& -  \KL{P_{\vect{z}_2}}{P_{S}} + \KL{P_{\vect{z}_1}}{P_{S}}\Bigl),
	\end{IEEEeqnarray}
where the function~$\mathsf{L}$ is in~\eqref{EqLxy}; the model~$\vect{\theta} \in \set{M}$, and the measures~$P_S$ and~$\worstgeneral$ satisfy~\eqref{Eqtheworstgeneral}.
\end{theorem}
\begin{IEEEproof}
	The proof follows from the equality in~\eqref{definitionrecall},
which together with Theorem~\ref{theoremgengapexpression} completes the proof.
\end{IEEEproof}

In Theorem~\ref{theoremdifferenceL}, the reference measure~$P_S$ can be arbitrarily chosen  as long as both types~$P_{\vect{z}_1}$ and~$P_{\vect{z}_2}$ are absolutely continuous with~$P_S$.
A choice that satisfies this constraint is the type induced by the aggregation of both datasets~$\vect{z}_1$ and~$\vect{z}_2$, which is denoted by~$\vect{z}_{0} = \left( \vect{z}_{1}, \vect{z}_{2} \right) \in \left(\set{X} \times \set{Y}\right)^{n_0}$, with~$n_0 = n_1 + n_2$. The type induced by the aggregated dataset~$\vect{z}_{0}$, denoted by~$P_{\vect{z}_0}$,  is a convex combination of the types~$P_{\vect{z}_1}$ and~$P_{\vect{z}_2}$, that is,~$P_{\vect{z}_0} = \frac{n_1}{n_0}P_{\vect{z}_1} + \frac{n_2}{n_0}P_{\vect{z}_2}$, which satisfies the absolute continuity conditions \cite{Perlaza-ISIT2023b}. 

%

From Theorem~\ref{theoremdifferenceL}, it appears  that the difference between a test empirical risk $\mathsf{L}(\vect{z}_1,\vect{\theta}) $ and the training empirical risk $\mathsf{L}(\vect{z}_2,\vect{\theta})$ of a given model $\vect{\theta}$ is determined by two values: $(a)$ the difference of the ``statistical distance'' from the types induced by the training and test datasets to the worst-case data-generating probability measure, i.e., $\KL{P_{\vect{z}_2}}{\worstgeneral} - \KL{P_{\vect{z}_1}}{\worstgeneral}$; and $(b)$ the difference of the ``statistical distance'' from the types to the reference measure  $P_S$, i.e., $\KL{P_{\vect{z}_1}}{P_{S}} - \KL{P_{\vect{z}_2}}{P_{S}}$. 

 \section{Analysis of the Generalization Gap}\label{sectiongeneralizationgap}

The generalization gap induced by a given model~$\vect{\theta} \in \set{M}$, which is assumed to be obtained with a training dataset~$\vect{z} \in \left( \set{X} \times \set{Y} \right)^n$, under the assumption that training and test datasets are independent and identically distributed according to the probability measure~$P_{Z} \in \triangle \left( \set{X} \times \set{Y} \right)$, is
\begin{IEEEeqnarray}{rCl}
	\nonumber
	& &G(\vect{\theta}, P_Z, P_{\vect{z}}) \\
	\label{Eqgengaptype}
	& = & \int\loss\mathrm{d}P_{Z}(x,y)  -  \int\loss\mathrm{d}P_{\vect{z}}(x,y)\supersqueezeequ. \spnum
\end{IEEEeqnarray}
The term 	$\int\loss\mathrm{d}P_{\vect{z}}(x,y) = \mathsf{L}(\vect{z}, \vect{\theta})$ is  an empirical risk often referred to as the training risk or training loss \cite{shalev2014understanding}. This is essentially the loss induced by the model with respect to the dataset used for training.  
The term~$\int\loss\mathrm{d}P_{Z}(x,y)$ is the population risk, also known as true risk. That is, the expected loss under the assumption that the ground-truth probability distribution of the data points is~$P_Z$.
Interestingly, as shown in~\eqref{Eqgengaptype}, such  generalization error can be written in terms of the functional~$G$  in~\eqref{Eqgengap}. This observation leads to the following description of the generalization gap.
\begin{lemma}\label{LemmaWrittenInTheBus203A}
	The generalization gap~$G(\vect{\theta},P_Z,P_{\vect{z}})$ in~\eqref{Eqgengaptype} satisfies:
	\begin{IEEEeqnarray}{rCl}
		\nonumber
		&&G(\vect{\theta},P_Z,P_{\vect{z}}) = \\
\label{EqProbablyANiceInterpretation1}
		&& \beta\Bigl(\KL{P_{\vect{z}}}{\worstmeasure{P_Z}} - \KL{P_{\vect{z}}}{P_Z} 
		- \KL{P_Z}{\worstmeasure{P_Z}}\Bigl), \Tsupersqueezeequ \IEEEeqnarraynumspace
	\end{IEEEeqnarray}
	where the measure~$\worstmeasure{P_Z}$ is the solution to the optimization problem in~\eqref{theworstproblem} under the assumption that~$P_S = P_Z$. 
\end{lemma}
\begin{IEEEproof}
	The proof follows from Corollary~\ref{CorRGspecifica} by noticing that the type~$P_{\vect{z}}$ is absolutely continuous with respect to~$P_{Z}$. 
\end{IEEEproof}

Lemma~\ref{LemmaWrittenInTheBus203A} highlights the intuition that if the type~$P_{\vect{z}}$ induced by the training dataset~$\vect{z}$ is at an arbitrary small ``statistical distance'' of the ground-truth measure~$P_{Z}$, the generalization gap~$G(\vect{\theta},P_Z,P_{\vect{z}})$ in~\eqref{Eqgengaptype}  is arbitrarily close to zero. This is revealed by the fact that an arbitrary small value of~$\KL{P_{\vect{z}}}{P_Z}$ implies the difference~$\KL{P_{\vect{z}}}{\worstmeasure{P_Z}} - \KL{P_Z}{\worstmeasure{P_Z}}$ is also arbitrarily small.

A more general expression for the generalization gap~$G(\vect{\theta},P_Z,P_{\vect{z}})$ in~\eqref{Eqgengaptype}  is provided by the following corollary of Theorem~\ref{theoremgengapexpression}. 
\begin{corollary}\label{CorCloseToTheDeadlineAAAI2024}
	The generalization gap~$G(\vect{\theta},P_Z,P_{\vect{z}})$ in~\eqref{Eqgengaptype} satisfies:
	\begin{IEEEeqnarray}{rCl}
		\nonumber
		G(\vect{\theta},P_Z,P_{\vect{z}})
		&=& \beta\Bigl(\KL{P_{\vect{z}}}{\worstgeneral}-\KL{P_Z}{\worstgeneral} \\
		\label{EqProbablyANiceInterpretation2}
		&&- \KL{P_{\vect{z}}}{P_S} + \KL{P_Z}{P_S}\Bigl),\spnum
	\end{IEEEeqnarray}
	where the measure~$\worstmeasure{P_Z}$ is in~\eqref{theworstproblem}. 
\end{corollary}

Note that several expressions for the generalization gap $G(\vect{\theta},P_Z,P_{\vect{z}})$ in~\eqref{Eqgengaptype} can be obtained from Corollary~\ref{CorCloseToTheDeadlineAAAI2024} by choosing the reference~$P_{S}$ and the parameter~$\gamma$ in~\eqref{theworstproblem}, which determines the value of~$\beta$.

\subsection{Expected Generalization Gap}

A conditional probability distribuition~$P_{\vect{\Theta} | \vect{Z}}$, such that given a training dataset~$\vect{z} \in \left( \set{X} \times \set{Y} \right)^n$, the measure~$P_{\vect{\Theta} | \vect{Z} = \vect{z}} \in \Bormeaspace{\set{M}}$ is used to choose models, is referred to as a statistical learning algorithm. This subsection, provides explicit expressions for the generalization gap induced by the algorithm~$P_{\vect{\Theta} | \vect{Z}}$ and a given training dataset.

The generalization gap~$G(\vect{\theta},P_Z,P_{\vect{z}})$ in~\eqref{Eqgengaptype}  is due to a particular model~$\vect{\theta}$, which has been deterministically obtained  from the training dataset~$\vect{z}$. 
When the model is chosen by using a statistical learning algorithm~$P_{\vect{\Theta} | \vect{Z}}$, trained upon the dataset~$\vect{z}$, the expected generalization gap is the expectation of~$G(\vect{\theta},P_Z,P_{\vect{z}})$ when $\vect{\theta}$ is sampled from~$P_{\vect{\Theta} | \vect{Z} = \vect{z}}$. 
Let~$\overline{G}: \triangle\Bormeaspace{\set{M}} \times \triangle\left( \set{X} \times \set{Y} \right) \times \triangle\left( \set{X} \times \set{Y} \right) \to \reals$ be such that
\begin{IEEEeqnarray}{rcl}
	\label{EqBarG}
	\overline{G}(P_{\vect{\Theta} | \vect{Z} = \vect{z}},P_Z,P_{\vect{z}}) & = & \int G(\vect{\theta},P_Z,P_{\vect{z}}) \mathrm{d} P_{\vect{\Theta} | \vect{Z} = \vect{z}}\left( \vect{\theta} \right),\IEEEeqnarraynumspace
\end{IEEEeqnarray}
where the functional $G$ is in \eqref{Eqgengaptype}.
Using this notation, the expected generalization error induced by the algorithm~$P_{\vect{\Theta} | \vect{Z}}$, when the training dataset is~$\vect{z}$, is~$\overline{G}(P_{\vect{\Theta} | \vect{Z} = \vect{z}},P_Z,P_{\vect{z}})$ in~\eqref{EqBarG}.
Corollary~\ref{CorCloseToTheDeadlineAAAI2024}, by appropriately choosing the reference measure~$P_{S}$ and the parameter~$\gamma$ in~\eqref{theworstproblem}, leads to numerous closed-form expressions for the expected generalization gap induced by the algorithm~$P_{\vect{\Theta} | \vect{Z}}$ for the training dataset $\vect{z}$. Interestingly, regardless of the choice of~$P_S$ and~$\gamma$, the resulting expressions describe the impact of the training dataset~$\vect{z}$ on the expected generalization gap.

\subsection{Doubly-Expected Generalization Gap}

The expected generalization gap~$\overline{G}(P_{\vect{\Theta} | \vect{Z} = \vect{z}},P_Z,P_{\vect{z}})$ in~\eqref{EqBarG} depends on the training dataset~$\vect{z}$. The doubly-expected generalization gap is obtained by taking the expectation of~$\overline{G}(P_{\vect{\Theta} | \vect{Z} = \vect{z}},P_Z,P_{\vect{z}})$ when~$\vect{z} \in \left( \set{X} \times \set{Y} \right)^n$ is sampled from~$P_{\vect{Z}}$, which is assumed to be the product distribution formed by~$P_{Z}$. 
Let~$\overline{\overline{G}}: \triangle\Bigl( \set{M} | \left(\set{X} \times \set{Y} \right)^n \Bigr)\times \triangle \Bigl(\left(\set{X} \times \set{Y} \right)^n \Bigr)\to \reals$ be a functional such that
\begin{IEEEeqnarray}{rcl}
	\label{Eq2BarG}
	\overline{\overline{G}}(P_{\vect{\Theta} | \vect{Z}},P_Z ) & = & \int \int G(\vect{\theta},P_Z,P_{\vect{z}}) \mathrm{d} P_{\vect{\Theta} | \vect{Z} = \vect{z}}\left( \vect{\theta} \right) \mathrm{d} P_{\vect{Z}}\left( \vect{z}\right),\Dsupersqueezeequ \IEEEeqnarraynumspace
\end{IEEEeqnarray}
where the functional $G$ is in \eqref{Eqgengaptype}.
Using this notation, the doubly-expected generalization error induced by the algorithm~$P_{\vect{\Theta} | \vect{Z}}$ is~$\overline{\overline{G}}(P_{\vect{\Theta} | \vect{Z}},P_Z )$ in~\eqref{Eq2BarG}.
In existing literature, the doubly-expected generalization gap is simply referred to as generalization error. See for instance \cite{xu2017information}, \cite{aminian2021exact}, and  \cite{InriaRR9454}. Note that in these previous works, the dependence on a particular training dataset is not explicit due to results being presented for the case in which the expectation is taken with respect to all sources of randomness in the corresponding expression.
As in the case of the expected generalization gap, Corollary~\ref{CorCloseToTheDeadlineAAAI2024} leads to numerous closed-form expressions for the doubly-expected generalization gap induced by the algorithm~$P_{\vect{\Theta} | \vect{Z}}$. 

\subsection{The Gibbs Algorithm}

A typical statistical learning algorithm is the Gibbs algorithm, which is parametrized by a positive real~$\lambda$ and by a~$\sigma$-measure~$Q \in \triangle \Bormeaspace{\set{M}}$ \cite{InriaRR9454}. The probability measure representing such an algorithm,  which is denoted by~$P^{\left(Q, \lambda\right)}_{\vect{\Theta}| \vect{Z}}$,  satisfies for all~$\vect{\theta} \in \supp Q$ and for all~$\vect{z} \in \left(\set{X} \times\set{Y} \right)^n$,
\begin{IEEEeqnarray}{rcl}\label{EqGenpdf}
	\frac{\mathrm{d}P^{\left(Q, \lambda\right)}_{\vect{\Theta}| \vect{Z} = \vect{z}}}{\mathrm{d}Q} \left( \vect{\theta} \right) 
	& =& \exp\left( - K_{Q, \vect{z}}\left(-\frac{1}{\lambda} \right) - \frac{1}{\lambda} \mathsf{L}\left( \vect{z},\vect{\theta}\right)\right),\spnum
\end{IEEEeqnarray}
where  the dataset~$\vect{z}$ represents the training dataset; the function $\mathsf{L}$ is defined in \eqref{EqLxy}; and the function~$K_{Q, \vect{z}}: \reals \to \reals$ satisfies $K_{Q, \vect{z}}\left( t \right) = ~\log \left( \int \exp\left( t\; \mathsf{L}\left( \vect{z},\vect{\nu}\right)  \right) \mathrm{d}Q(\vect{\nu}) \right)$.

The doubly-expected generalization error induced by the Gibbs algorithm with parameters~$Q$ and~$\lambda$, under the assumption that datasets are sampled from a product distribution formed by the measure~$P_Z$,  denoted~$\overline{\overline{G}}(P^{\left(Q, \lambda\right)}_{\vect{\Theta}| \vect{Z}},P_Z )$ satisfies the following property.
\begin{lemma}[Generalization Gap of the Gibbs Algorithm]\label{LemmaOneMoreProofGenError}
	Given the conditional probability measure~$P^{\left(Q, \lambda\right)}_{\vect{\Theta}| \vect{Z}}$ in~\eqref{EqGenpdf} and a probability measure~$P_Z \in \triangle\left( \set{X} \times \set{Y} \right)$, 
	the generalization gap~$\overline{\overline{G}}(P^{\left(Q, \lambda\right)}_{\vect{\Theta}| \vect{Z}},P_Z )$ satisfies
	\begin{IEEEeqnarray}{rcl}
		\overline{\overline{G}}(P^{\left(Q, \lambda\right)}_{\vect{\Theta}| \vect{Z}},P_Z ) & = &  \lambda \left( I\left( P^{\left(Q, \lambda\right)}_{\vect{\Theta}| \vect{Z}}; P_{\vect{Z}} \right) +  L\left( P^{\left(Q, \lambda\right)}_{\vect{\Theta}| \vect{Z}}; P_{\vect{Z}} \right) \right),\Dsupersqueezeequ \IEEEeqnarraynumspace
	\end{IEEEeqnarray}
	where~$P_{\vect{Z}} \in \triangle\left(\set{X} \times\set{Y} \right)^n$ is a product measure obtained from~$P_{Z}$; and~$I\left( P^{\left(Q, \lambda\right)}_{\vect{\Theta}| \vect{Z}}; P_{\vect{Z}} \right)$ and~$L\left( P^{\left(Q, \lambda\right)}_{\vect{\Theta}| \vect{Z}}; P_{\vect{Z}} \right)$ are, respectively, the mutual information and the lautum information given by
	\begin{IEEEeqnarray}{rcl}
		\label{mutualdef}
		I\left( P^{\left(Q, \lambda\right)}_{\vect{\Theta}| \vect{Z}}; P_{\vect{Z}} \right) & \triangleq & 
		\int  \KL{P^{\left(Q, \lambda\right)}_{\vect{\Theta}| \vect{Z} = \vect{\nu}} }{P^{\left(Q, \lambda\right)}_{\vect{\Theta}}}  \mathrm{d} P_{\vect{Z}}(\vect{\nu}); \mbox{ and }  \Dsupersqueezeequ \IEEEeqnarraynumspace\\
		\label{EqPulgosos}
		L\left( P^{\left(Q, \lambda\right)}_{\vect{\Theta}| \vect{Z}}; P_{\vect{Z}} \right) & \triangleq & 
		\int \KL{P^{\left(Q, \lambda\right)}_{\vect{\Theta}}}{P^{\left(Q, \lambda\right)}_{\vect{\Theta}| \vect{Z} = \vect{\nu}} }  \mathrm{d} P_{\vect{Z}}(\vect{\nu})   ,\Dsupersqueezeequ \IEEEeqnarraynumspace
	\end{IEEEeqnarray}
	with the probability measure~$P^{\left(Q, \lambda\right)}_{\vect{\Theta}}$ being such that for all sets~$\set{A} \in \BorSigma{\set{M}}$,	\label{EqBarPThetaX}
		$P^{\left(Q, \lambda\right)}_{\vect{\Theta}}\left( \set{A} \right) = \int P^{\left(Q, \lambda\right)}_{\vect{\Theta} | \vect{Z} = \vect{\nu}  } \left( \set{A} \right)  \mathrm{d}P_{\vect{Z}}\left( \vect{\nu} \right)$.
\end{lemma}
\begin{IEEEproof}
	The proof is presented in \cite[Appendix H]{InriaRR9515}.
\end{IEEEproof}
Lemma~\ref{LemmaOneMoreProofGenError} has been proved before for the case in which~$Q$ is a probability measure in \cite{aminian2021exact}; and in the more general case in which~$Q$ is a~$\sigma$-finite measure in \cite{InriaRR9454}. In both \cite{aminian2021exact} and \cite{InriaRR9454}, the result is shown without the assumption that the measure~$P_{\vect{Z}}$ is a product measure, which is an assumption in Lemma~\ref{LemmaOneMoreProofGenError}. This limitation is due to the fact that the proof of  Lemma~\ref{LemmaOneMoreProofGenError} relies on the notion of types, which is known to fail capturing the correlation between datapoints, as pointed in \cite{csiszar1998method}.  Nonetheless, the independent and identically distributed assumption is widely adopted in the realm of machine learning.
Despite this limitation, the relevance of  Lemma~\ref{LemmaOneMoreProofGenError} stems from the fact that a connection has been made between the notion of sensitivity to deviations from the worst-case data-generating measure, which is captured by the functional~$G$ in~\eqref{Eqgengap}, and the notion of (doubly-expected) generalization gap, which is a central performance metric for evaluating the generalization capabilities of machine learning algorithms.  

\section{Conclusions and Final Remarks} 
 
The worst-case data-generating probability measure in Theorem~\ref{theworstlemma} has been shown to be a cornerstone in statistical machine learning. This is due to the fact that  fundamental performance metrics, such as the sensitivity of the expected loss, the sensitivity of the empirical risk, the expected generalization gap, and the doubly-expected generalization gap are shown to have closed-form expressions  involving such a measure. 
The dependence of these performance metrics on the worst-case data-generating probability measure is shown to exist via the sensitivity of the expectation of the loss function to changes from the worst-case data-generating probability measure to any alternative probability measure.
This observation is reminiscent of the dependence of the expected generalization gap and the doubly-expected generalization gap on a Gibbs probability measure on the measurable space of the models as shown in \cite{InriaRR9454, Perlaza-ISIT-2022, InriaRR9474}.  
%
%
These dependences suggest an intriguing relation between the probability measure (on the models) describing the Gibbs algorithm and the worst-case probability measure (on the datasets) introduced in this work, which is also a Gibbs probability measure. The connection appears to be nontrivial and is suggested as a promising line of work in this area.

\section{Acknowledgments}

This work is funded in part by the ANR Project PARFAIT under grant ANR-21-CE25-0013 and the INRIA Exploratory Action IDEM.

\bibliography{reference}

\begin{thebibliography}{36}
\providecommand{\natexlab}[1]{#1}

\bibitem[{Aminian et~al.(2021)Aminian, Bu, Toni, Rodrigues, and
  Wornell}]{aminian2021exact}
Aminian, G.; Bu, Y.; Toni, L.; Rodrigues, M.; and Wornell, G. 2021.
\newblock An Exact Characterization of the Generalization Error for the {G}ibbs
  Algorithm.
\newblock \emph{Advances in Neural Information Processing Systems}, 34:
  8106--8118.

\bibitem[{Aminian et~al.(2022)Aminian, Bu, Wornell, and
  Rodrigues}]{aminian2022tighter}
Aminian, G.; Bu, Y.; Wornell, G.~W.; and Rodrigues, M.~R. 2022.
\newblock Tighter expected generalization error bounds via convexity of
  information measures.
\newblock In \emph{Proceedings of the IEEE International Symposium on
  Information Theory (ISIT)}, 2481--2486. Aalto, Finland.

\bibitem[{Aminian, Toni, and Rodrigues(2021)}]{aminian2021jensen}
Aminian, G.; Toni, L.; and Rodrigues, M.~R. 2021.
\newblock {J}ensen-{S}hannon information based characterization of the
  generalization error of learning algorithms.
\newblock In \emph{Proceedings of the IEEE Information Theory Workshop (ITW)},
  1--5. Kanazawa, Japan.

\bibitem[{Asadi, Abbe, and Verd{\'u}(2018)}]{asadi2018chaining}
Asadi, A.; Abbe, E.; and Verd{\'u}, S. 2018.
\newblock Chaining mutual information and tightening generalization bounds.
\newblock \emph{Advances in Neural Information Processing Systems}, 31:
  7245--7254.

\bibitem[{Asadi and Abbe(2020)}]{asadi2020chaining}
Asadi, A.~R.; and Abbe, E. 2020.
\newblock Chaining Meets Chain Rule: {M}ultilevel Entropic Regularization and
  Training of Neural Networks.
\newblock \emph{The Journal of Machine Learning Research}, 21(1): 5453--5484.

\bibitem[{Bu, Zou, and Veeravalli(2020)}]{bu2020tightening}
Bu, Y.; Zou, S.; and Veeravalli, V.~V. 2020.
\newblock Tightening mutual information-based bounds on generalization error.
\newblock \emph{IEEE Journal on Selected Areas in Information Theory}, 1(1):
  121--130.

\bibitem[{Cherkassky et~al.(1999)Cherkassky, Shao, Mulier, and
  Vapnik}]{cherkassky1999model}
Cherkassky, V.; Shao, X.; Mulier, F.~M.; and Vapnik, V.~N. 1999.
\newblock Model complexity control for regression using {VC} generalization
  bounds.
\newblock \emph{IEEE Transactions on Neural Networks}, 10(5): 1075--1089.

\bibitem[{Chu and Raginsky(2023)}]{chu2023unified}
Chu, Y.; and Raginsky, M. 2023.
\newblock A unified framework for information-theoretic generalization bounds.
\newblock \emph{\normalfont{arXiv preprint arXiv:2305.11042}}.

\bibitem[{Csisz{\'a}r(1998)}]{csiszar1998method}
Csisz{\'a}r, I. 1998.
\newblock The method of types.
\newblock \emph{IEEE Transactions on Information Theory}, 44(6): 2505--2523.

\bibitem[{Cullina, Bhagoji, and Mittal(2018)}]{cullina2018pac}
Cullina, D.; Bhagoji, A.~N.; and Mittal, P. 2018.
\newblock {PAC}-learning in the presence of adversaries.
\newblock \emph{Advances in Neural Information Processing Systems}, 31(1):
  1--12.

\bibitem[{Daunas et~al.(2023{\natexlab{a}})Daunas, Esnaola, Perlaza, and
  Poor}]{Perlaza-ISIT2023a}
Daunas, F.; Esnaola, I.; Perlaza, S.~M.; and Poor, H.~V. 2023{\natexlab{a}}.
\newblock Analysis of the Relative Entropy Asymmetry in the Regularization of
  Empirical Risk Minimization.
\newblock In \emph{Proceedings of the IEEE International Symposium on
  Information Theory (ISIT)}, 340--345. Taipei, Taiwan.

\bibitem[{Daunas et~al.(2023{\natexlab{b}})Daunas, Esnaola, Perlaza, and
  Poor}]{InriaRR9521}
Daunas, F.; Esnaola, I.; Perlaza, S.~M.; and Poor, H.~V. 2023{\natexlab{b}}.
\newblock Empirical Risk Minimization with f-Divergence Regularization in
  Statistical Learning.
\newblock Technical Report RR-9521, INRIA, Centre Inria d'Universit\'e C\^ote
  d'Azur, Sophia Antipolis, France.

\bibitem[{Dembo and Zeitouni(2009)}]{dembo2009large}
Dembo, A.; and Zeitouni, O. 2009.
\newblock \emph{Large Deviations Techniques and Applications}.
\newblock New York, NY, USA: Springer-Verlag, 2nd edition.

\bibitem[{Duchi, Glynn, and Namkoong(2021)}]{duchi2021statistics}
Duchi, J.~C.; Glynn, P.~W.; and Namkoong, H. 2021.
\newblock Statistics of robust optimization: A generalized empirical likelihood
  approach.
\newblock \emph{Mathematics of Operations Research}, 46(3): 946--969.

\bibitem[{Esposito, Gastpar, and Issa(2020)}]{esposito2020robust}
Esposito, A.~R.; Gastpar, M.; and Issa, I. 2020.
\newblock Robust Generalization via $\alpha$-Mutual Information.
\newblock \emph{\normalfont{arXiv preprint arXiv:2001.06399}}.

\bibitem[{Georgii(2011)}]{georgii2011gibbs}
Georgii, H.-O. 2011.
\newblock \emph{{G}ibbs measures and phase transitions}.
\newblock New York, NY, USA: De Gruyter, 2nd edition.

\bibitem[{Haddouche et~al.(2021)Haddouche, Guedj, Rivasplata, and
  Shawe-Taylor}]{haddouche2020pacbayes}
Haddouche, M.; Guedj, B.; Rivasplata, O.; and Shawe-Taylor, J. 2021.
\newblock {PAC-Bayes} unleashed: {G}eneralisation bounds with unbounded losses.
\newblock \emph{Entropy}, 23(10): 1--20.

\bibitem[{Hafez-Kolahi et~al.(2020)Hafez-Kolahi, Golgooni, Kasaei, and
  Soleymani}]{hafez2020conditioning}
Hafez-Kolahi, H.; Golgooni, Z.; Kasaei, S.; and Soleymani, M. 2020.
\newblock Conditioning and processing: {T}echniques to improve
  information-theoretic generalization bounds.
\newblock \emph{Advances in Neural Information Processing Systems},
  16457--16467.

\bibitem[{Hellstr{\"o}m and Durisi(2020)}]{hellstrom2020generalization}
Hellstr{\"o}m, F.; and Durisi, G. 2020.
\newblock Generalization bounds via information density and conditional
  information density.
\newblock \emph{IEEE Journal on Selected Areas in Information Theory}, 1(3):
  824--839.

\bibitem[{Issa, Esposito, and Gastpar(2019)}]{issa2019strengthened}
Issa, I.; Esposito, A.~R.; and Gastpar, M. 2019.
\newblock Strengthened information-theoretic bounds on the generalization
  error.
\newblock In \emph{Proceedings of the IEEE International Symposium on
  Information Theory (ISIT)}, 582--586. Paris, France.

\bibitem[{Jeffreys(1946)}]{jeffreys1946invariant}
Jeffreys, H. 1946.
\newblock An invariant form for the prior probability in estimation problems.
\newblock \emph{Proceedings of the Royal Society of London. {Series A}.
  Mathematical and Physical Sciences}, 186(1007): 453--461.

\bibitem[{Lee and Raginsky(2018)}]{lee2018minimax}
Lee, J.; and Raginsky, M. 2018.
\newblock Minimax statistical learning with Wasserstein distances.
\newblock \emph{Advances in Neural Information Processing Systems}, 31:
  2687--2696.

\bibitem[{Lopez and Jog(2018)}]{lopez2018generalization}
Lopez, A.~T.; and Jog, V. 2018.
\newblock Generalization error bounds using Wasserstein distances.
\newblock In \emph{Proceedings of the IEEE Information Theory Workshop (ITW)},
  1--5. Guangzhou, China.

\bibitem[{Masiha, Gohari, and Yassaee(2023)}]{masiha2023f}
Masiha, S.; Gohari, A.; and Yassaee, M.~H. 2023.
\newblock f-divergences and their applications in lossy compression and
  bounding generalization error.
\newblock \emph{IEEE Transactions on Information Theory}, 69(12): 7245--7254.

\bibitem[{Mazuelas, Shen, and P{\'e}rez(2022)}]{mazuelas2022generalized}
Mazuelas, S.; Shen, Y.; and P{\'e}rez, A. 2022.
\newblock Generalized maximum entropy for supervised classification.
\newblock \emph{IEEE Transactions on Information Theory}, 68(4): 2530--2550.

\bibitem[{McAllester(2003)}]{mcallester2003pac}
McAllester, D.~A. 2003.
\newblock {PAC-B}ayesian stochastic model selection.
\newblock \emph{Machine Learning}, 51(1): 5--21.

\bibitem[{Perlaza et~al.(2022{\natexlab{a}})Perlaza, Bisson, Esnaola,
  Jean-Marie, and Rini}]{InriaRR9454}
Perlaza, S.~M.; Bisson, G.; Esnaola, I.; Jean-Marie, A.; and Rini, S.
  2022{\natexlab{a}}.
\newblock Empirical Risk Minimization with Relative Entropy Regularization.
\newblock Technical Report RR-9454, INRIA, Centre Inria d'Universit\'e C\^ote
  d'Azur, Sophia Antipolis, France.

\bibitem[{Perlaza et~al.(2022{\natexlab{b}})Perlaza, Bisson, Esnaola,
  Jean-Marie, and Rini}]{Perlaza-ISIT-2022}
Perlaza, S.~M.; Bisson, G.; Esnaola, I.; Jean-Marie, A.; and Rini, S.
  2022{\natexlab{b}}.
\newblock Empirical Risk Minimization with Relative Entropy Regularization:
  {O}ptimality and Sensitivity.
\newblock In \emph{Proceedings of the IEEE International Symposium on
  Information Theory (ISIT)}, 684--689. Espoo, Finland.

\bibitem[{Perlaza et~al.(2022{\natexlab{c}})Perlaza, Esnaola, Bisson, and
  Poor}]{InriaRR9474}
Perlaza, S.~M.; Esnaola, I.; Bisson, G.; and Poor, H.~V. 2022{\natexlab{c}}.
\newblock Sensitivity of the {G}ibbs Algorithm to Data Aggregation in
  Supervised Machine Learning.
\newblock Technical Report RR-9474, INRIA, Centre Inria d'Universit\'e C\^ote
  d'Azur, Sophia Antipolis, France.

\bibitem[{Perlaza et~al.(2023)Perlaza, Esnaola, Bisson, and
  Poor}]{Perlaza-ISIT2023b}
Perlaza, S.~M.; Esnaola, I.; Bisson, G.; and Poor, H.~V. 2023.
\newblock On the Validation of {G}ibbs Algorithms: {T}raining Datasets, Test
  Datasets and their Aggregation.
\newblock In \emph{Proceedings of the IEEE International Symposium on
  Information Theory (ISIT)}, 328--333. Taipei, Taiwan.

\bibitem[{Russo and Zou(2016)}]{pmlr-v51-russo16}
Russo, D.; and Zou, J. 2016.
\newblock Controlling Bias in Adaptive Data Analysis Using Information Theory.
\newblock In \emph{Proceedings of the 19th International Conference on
  Artificial Intelligence and Statistics}, volume~51, 1232--1240. Cadiz, Spain.

\bibitem[{Russo and Zou(2019)}]{russo2019much}
Russo, D.; and Zou, J. 2019.
\newblock How much does your data exploration overfit? {C}ontrolling bias via
  information usage.
\newblock \emph{IEEE Transactions on Information Theory}, 66(1): 302--323.

\bibitem[{Shalev-Shwartz and Ben-David(2014)}]{shalev2014understanding}
Shalev-Shwartz, S.; and Ben-David, S. 2014.
\newblock \emph{Understanding Machine Learning: {F}rom {T}heory to
  {A}lgorithms}.
\newblock New York, NY, USA: Cambridge University Press, 1st edition.

\bibitem[{Wang et~al.(2019)Wang, Diaz, Santos~Filho, and
  Calmon}]{wang2019information}
Wang, H.; Diaz, M.; Santos~Filho, J. C.~S.; and Calmon, F.~P. 2019.
\newblock An information-theoretic view of generalization via {W}asserstein
  distance.
\newblock In \emph{Proceedings of the IEEE International Symposium on
  Information Theory (ISIT)}, 577--581. Paris, France.

\bibitem[{Xu and Raginsky(2017)}]{xu2017information}
Xu, A.; and Raginsky, M. 2017.
\newblock Information-theoretic analysis of generalization capability of
  learning algorithms.
\newblock \emph{Advances in Neural Information Processing Systems}, 30: 1--10.

\bibitem[{Zou et~al.(2023)Zou, Perlaza, Esnaola, and Altman}]{InriaRR9515}
Zou, X.; Perlaza, S.~M.; Esnaola, I.; and Altman, E. 2023.
\newblock The {W}orst-{C}ase {D}ata-{G}enerating {P}robability {M}easure.
\newblock Technical Report RR-9515, INRIA, Centre Inria d'Universit\'e C\^ote
  d'Azur, Sophia Antipolis, France.

\end{thebibliography}

\end{document}